\documentclass[10pt, a4paper]{article}
\usepackage{lrec2022} 
\usepackage{multibib}
\newcites{languageresource}{Language Resources}
\usepackage{graphicx}
\usepackage{tabularx}
\usepackage{float}
\usepackage{soul}
\usepackage[T1]{fontenc} 
\usepackage[utf8]{inputenc}
\usepackage{enumitem}
\usepackage{titlesec}
\titleformat{\section}{\normalfont\large\bfseries\center}{\thesection.}{1em}{}
\titleformat{\subsection}{\normalfont\SmallTitleFont\bfseries\raggedright}{\thesubsection.}{1em}{}
\titleformat{\subsubsection}{\normalfont\normalsize\bfseries\raggedright}{\thesubsubsection.}{1em}{}
\renewcommand\thesection{\arabic{section}}
\renewcommand\thesubsection{\thesection.\arabic{subsection}}
\renewcommand\thesubsubsection{\thesubsection.\arabic{subsubsection}}

\usepackage{ifpdf}
\ifpdf
  \usepackage{epstopdf}
\fi

\usepackage{hyperref}
\usepackage{xstring}
\usepackage{color}
\usepackage{tikz}
\usepackage{array,multirow,graphicx}
\usetikzlibrary{shapes.geometric, arrows}

\title{Introducing the Welsh Text Summarisation Dataset and Baseline Systems}

\name{Ignatius Ezeani$^{1}$, Mahmoud El-Haj$^{1}$, Jonathan Morris$^{2}$ and Dawn Knight$^{2}$}

\address{$^{1}$UCREL NLP group, School of Computing and Communications, Lancaster University, $^{2}$Cardiff University\\
         \{i.ezeani, m.el-haj\}@lancaster.ac.uk, \{knightd5, morrisj17\}@cardiff.ac.uk\\
  }

\abstract{
Welsh is an official language in Wales and is spoken by an estimated 884,300 people (29.2\% of the population of Wales). Despite this status and estimated increase in speaker numbers since the last (2011) census, Welsh remains a minority language undergoing revitalisation and promotion by Welsh Government and relevant stakeholders. As part of the effort to increase the availability of Welsh digital technology, this paper introduces the first Welsh summarisation dataset, which we provide freely for research purposes to help advance the work on Welsh text summarisation. The dataset was created by Welsh speakers through manually summarising Welsh Wikipedia articles. In addition, the paper discusses the implementation and evaluation of different summarisation systems for Welsh. The summarisation systems and results will serve as benchmarks for the development of summarisers in other minority language contexts.
\newline \Keywords{summarisation, Welsh, corpus, word embeddings}}

\begin{document}

\maketitleabstract
\section{Introduction}\label{intro}
It is estimated that over a quarter (29.2\%) of the population in Wales aged over 3 consider themselves to be Welsh speakers \footnote{https://gov.wales/welsh-language-data-annual-population-survey-july-2020-june-2021}. Although this estimate represents an increase in the proportion of the population who reported speaking Welsh at the last (2011) census\footnote{https://statswales.gov.wales/Catalogue/Welsh-Language/Census-Welsh-Language}, historically the language has been in decline and represents a minority language in Wales despite having official status. This decline has led to the development of language policy designed to safeguard the language and promote its use among the population \cite{carlin2016standard}.

The most recent Welsh Government strategy for the revitalisation of Welsh has infrastructure (and particularly digital infrastructure) as a main theme along with increasing the number of speakers and increasing language use\footnote{Welsh Government: Cymraeg 2050 - A million Welsh speakers: \url{https://gov.wales/sites/default/files/publications/2018-12/cymraeg-2050-welsh-language-strategy.pdf}}. The aim is to ensure that the Welsh language is at the heart of innovation in digital technology to enable the use of Welsh in all digital contexts (Welsh Government 2017: 71). A system that could assist in the automatic summarisation of long documents would prove beneficial to the culture revitalisation efforts currently taking place. 

Over time, there have been various approaches to automatic text summarisation, but when looking at those approaches in detail, we can see that they are mainly split between \textit{single-document summarisation} (finding the most informative sentences in a document) and \textit{multi-document summarisation} (finding a summary that combines the main themes across thematically diverse set of documents) with the majority of work being applied to the English language, as a global lingua franca \cite{goldstein2000multi,svore2007enhancing,svore2007enhancing,litvak2008graph,el2011multi,el2013using}. \\ \\In this project, we focused on creating a high quality Welsh summarisation dataset containing entries similar to the sample shown in the example in Table \ref{tab:eg-text-summary-cy-en}. We went further to build and evaluate baseline systems that can produce summaries from single documents using basic extractive methods. The dataset and code for experiments and testing are available on the \href{https://github.com/UCREL/welsh-summarization-dataset}{Welsh Summarisation Project} GitHub page\footnote{https://github.com/UCREL/welsh-summarization-dataset} as well as the \href{https://share.streamlit.io/ucrel/welsh_summarizer/main/app/app.py}{application demo}\footnote{\url{https://share.streamlit.io/ucrel/welsh_summarizer/main/app/app.py}}.

\begin{table*}[ht]
\centering
\begin{tabular}{|p{7.3cm}||p{7.3cm}|}
\hline
\textbf{Welsh Text:} \textit{\hl{Mae} Erthygl 25 o \hl{Ddatganiad Cyffredinol Hawliau Dynol 1948} y Cenhedloedd Unedig \hl{yn nodi: "Mae gan bawb yr hawl i safon byw sy'n ddigonol} ar gyfer iechyd a lles ei hun a'i deulu, \hl{gan gynnwys bwyd, dillad}, tai a \hl{gofal meddygol} a gwasanaethau cymdeithasol angenrheidiol". Mae'r Datganiad Cyffredinol yn cynnwys lletyaeth er mwyn diogelu person ac mae hefyd yn sôn yn arbennig am y gofal a roddir i'r rheini sydd mewn mamolaeth neu blentyndod. Ystyrir y Datganiad Cyffredinol o Hawliau Dynol fel y \hl{datganiad} rhyngwladol \hl{cyntaf o hawliau dynol} sylfaenol. Dywedodd Uchel Gomisiynydd y Cenhedloedd Unedig dros Hawliau Dynol fod y Datganiad Cyffredinol o Hawliau Dynol yn ymgorffori gweledigaeth sy'n cynnwys yr holl hawliau dynol, sef hawliau sifil, gwleidyddol, economaidd, cymdeithasol neu ddiwylliannol.} &
\textbf{English Text:} \textit{Article 25 of \hl{the 1948 Universal Declaration of Human Rights} of the United Nations \hl{states: ``Everyone has the right to an adequate standard of living} for the health and well-being of himself and his family, \hl{including food, clothing}, housing and \hl{medical care}. and necessary social services". The General Statement includes accommodation to protect a person and also mentions the care given to those in maternity or childhood. The Universal Declaration of Human Rights is regarded as \hl{the first} international \hl{declaration of} basic \hl{human rights}. The United Nations High Commissioner for Human Rights said that the Universal Declaration of Human Rights embodies a vision that encompasses all human rights, civil, political, economic, social or cultural.}\\
\hline
\textbf{Welsh Summary:}
\textit{\hl{Mae Datganiad Cyffredinol Hawliau Dynol 1948 yn dweud bod gan bawb yr hawl i safon byw digonol. Mae hynny yn cynnwys mynediad at fwyd a dillad a gofal meddygol} i bob unigolyn. Dyma’r \hl{datganiad cyntaf o hawliau dynol.}}
&
\textbf{English Summary:}
\textit{\hl{The 1948 Universal Declaration of Human Rights states} that \hl{everyone has the right to an adequate standard of living}. This \hl{includes access to food and clothing and medical care} for each individual. This is \hl{the first declaration of human rights.}}\\
\hline
\end{tabular}
\caption{Example texts with human reference summaries in Welsh and English. System outputs are included in the Appendix}
\label{tab:eg-text-summary-cy-en}
\end{table*}

\section{Related Work}\label{relwork}
There exists a relatively low use of Welsh language websites and e-services, despite the fact that numerous surveys suggest that Welsh speakers would like more opportunities to use the language, and that there has been an expansive history of civil disobedience in order to gain language rights in the Welsh language context \cite{cunliffe2013young}. 
One reason for the relatively low take-up of Welsh-language options on websites is the assumption that the language used in such resources will be too complicated \cite{cunliffe2013young}. Concerns around the complexity of public-facing Welsh language services and documents are not new. A series of guidelines on creating easy-to-read documents in Welsh are outlined in Cymraeg Clir \cite{arthur2019human}. \newcite{Williams1999} notes that the need for simplified versions of Welsh is arguably greater than for English considering (1) many Welsh public-facing documents are translated from English, (2) the standard varieties of Welsh are further removed from local dialects compared to English, and (3) newly-translated technical terms are more likely to be familiar to the reader. The principles outlined in Cymraeg Clir therefore include the use of shorter sentences, everyday words rather than specialised terminology, and a neutral (rather than formal) register \cite{Williams1999}.
\\ \\
This paper reports on work on a project which aims to develop an online Automatic Text Summarisation tools for the Welsh language, ACC (Adnodd Creu Crynodebau).
ACC will provide the means for summarising and simplifying digital language sources, which will help in addressing the fears of Welsh speakers that language online is too complicated.
ACC will also contribute to the digital infrastructure of the Welsh language. Given the introduction of Welsh Language Standards \cite{carlin2016standard} and a concerted effort to both invest in Welsh language technologies and improve the way in which language choice is presented to the public, the development of ACC will complement the suite of Welsh language technologies (e.g. Canolfan Bedwyr 2021\footnote{Cysgliad: Help i ysgrifennu yn Gymraeg. Online: \url{https://www.cysgliad.com/cy/}}) for both content creators and Welsh readers. It is also envisaged that ACC will contribute to Welsh-medium education by allowing educators to create summaries for use in the classroom as pedagogical tools. Summaries will also be of use to Welsh learners who will be able to focus on understanding the key information within a text.

\section{Methods} \label{sec:methods}
Figure \ref{fig:process-diagram} shows the four key processes involved in the creation and testing of the Welsh summarisation dataset i.e. \textbf{a.} collection of the text data; \textbf{b.} creation of the reference (human) summaries; \textbf{c.} building summarisers and generating system summaries and \textbf{d.} evaluating the performance of the summarisation systems outputs on the reference summaries.

\subsection{Text Collection}
The first stage of the development process is to develop a small corpus (dataset) of target language data that will subsequently be summarised and evaluated by human annotators and used to develop and train the automated summarisation models (i.e. acting as a `gold-standard' dataset). 

\tikzstyle{startstop} = [rectangle, rounded corners, minimum width=3cm, minimum height=1cm, text centered, draw=black, fill=red!30]

\tikzstyle{process} = [rectangle, minimum width=3cm, minimum height=.75cm, text centered, draw=black, fill=orange!30]
\tikzstyle{arrow} = [thick,->,>=stealth]

\begin{figure}[h]
\centering
\begin{tikzpicture}[node distance=1.5cm]
\node (text-collection)[process]{Text Collection};
\node (text-cleaning)[process, below of=text-collection]{Text Cleaning};
\node (systems) [startstop, below of=text-cleaning, minimum height=4cm, minimum width=3.5cm,
yshift=-1cm, xshift=2cm] {};
\node (reference) [startstop, below of=text-cleaning, minimum height=2cm, minimum width=3.5cm,
yshift=-.5cm, xshift=-2cm] {};
\node (wiki_ref)[startstop, below of=text-cleaning, fill=yellow!30, minimum height=.75,
xshift=-2cm]{Wiki References};
\node (human_ref)[startstop, below of=wiki_ref, fill=yellow!30, minimum height=.75,
yshift=.5cm]{Human References};
\node (firstSent)[rectangle, above of=systems, fill=green!20, draw=black, minimum height=.75,
xshift=-.6cm, yshift=.2cm]{First Sent};
\node (TextRank)[rectangle, below of=firstSent, fill=green!20, draw=black, minimum height=.75,
xshift=.5cm, yshift=1cm]{TextRank};
\node (LexRank)[rectangle, below of=TextRank, fill=green!20, draw=black, minimum height=.75,
xshift=.5cm, yshift=1cm]{LexRank};
\node (TFIDF)[rectangle, below of=LexRank, fill=blue!15, draw=black, minimum height=.75,
xshift=-1.2cm, yshift=.85cm]{TFIDF};
\node (TFIDF-FastText)[rectangle, below of=TFIDF, fill=blue!15, draw=black, minimum height=.75,
xshift=1cm, yshift=1cm]{TFIDF-FastText};
\node (TFIDF-WNLT)[rectangle, below of=TFIDF-FastText, fill=blue!15, draw=black, minimum height=.75, xshift= .25cm, yshift=1cm]{TFIDF-WNLT};
\node (sys_ref)[startstop, below of=TFIDF-WNLT, fill=yellow!30, minimum height=.75,
xshift=-.45cm, yshift=.85cm]{System References};

\node (evaluation)[process, yshift=-7cm]{System Evaluation};

\draw [arrow] (text-collection) -- (text-cleaning);
\draw [arrow] (text-cleaning) -| (systems);
\draw [arrow] (text-cleaning) -| (reference);
\draw [arrow] (systems) |- (evaluation);
\draw [arrow] (reference) |- (evaluation);
\end{tikzpicture}
\caption{An overview of the process diagram with the key processes undertaken in this work. The components and processes are described and explained in Section \ref{sec:methods}.}
\label{fig:process-diagram}
\end{figure}
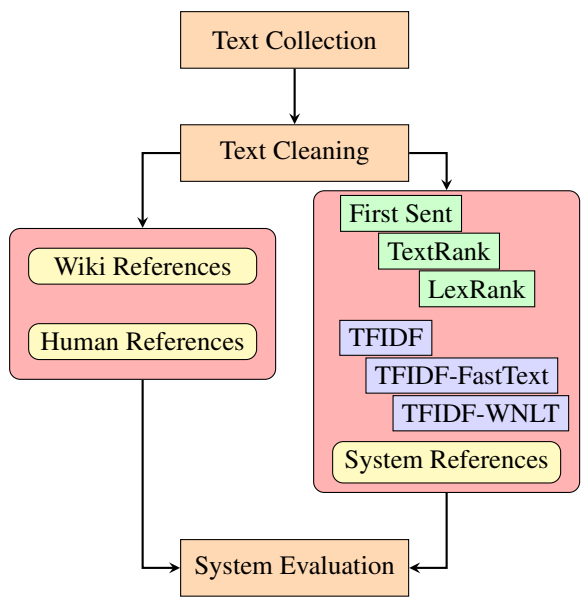

\begin{figure}[h]
\centering
\includegraphics[scale=0.55]{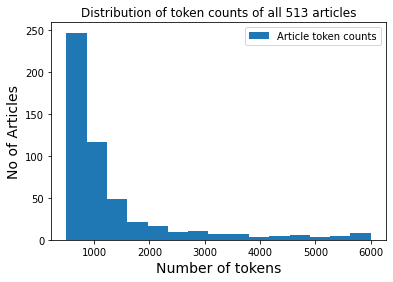}
\caption{Token counts of the 513 Wikipedia articles used for training of system summarisers as well as the average counts of the articles and the summaries. Majority of the articles (about 80\%) contain between 500 and 2000 tokens. A total of 28 articles contain more than 5000 tokens}
\label{fig:token-counts}
\end{figure}

Wikipedia\footnote{Welsh Wikipedia: \url{https://cy.wikipedia.org/wiki/Hafan} (Wicipedia)} was selected as the primary source of data for creating the Welsh language dataset for ACC. This was owing to the fact that an extensive number of Welsh language texts exist on this website (over 133,000 articles), all of which are available under GNU Free Documentation license. To ensure that pages that contained a sufficient quantity of text were extracted for use, a minimum threshold of 500 tokens per article and a target of at least 500 articles was established at the outset. A selection of 800 most accessed Wikipedia pages in Welsh were initially extracted for use. An additional 100 Wikipedia pages were included from the WiciAddysg project organised by the National Library of Wales and Menter Iaith Môn\footnote{WikiAddysg: \url{https://cy.wikipedia.org/wiki/Categori:Prosiect\_WiciAddysg}}. However, it was observed that more than 50\% of the articles from this original list of Wikipedia pages did not meet the minimum-token threshold of 500. To mitigate this, a list of 20 Welsh keywords was used to locate an additional 100 Wikipedia pages per keyword (which was provided by the third author, who is a native Welsh speaker, and contained words synonymous with the Welsh language, Welsh history and geography). This was added to the list of 100 most-edited Welsh Wikipedia pages and pages from the WiciAddysg project.
The data extraction applied a simple iterative process and implemented a Python script based on the WikipediaAPI\footnote{https://pypi.org/project/Wikipedia-API/} that takes a Wikipedia page; extracts key contents (article text, summary, category) and checks whether the article text contains a minimum number of tokens. At the end of this process, the dataset was created from a total of 513 Wikipedia pages that met the set criteria. Figure \ref{fig:token-counts} shows the distribution of the token counts for the 513 Wikipedia articles. The extracted dataset contains a file for each Wikipedia page with the following structure and tags:
\begin{center}
\small{\texttt{ <title>Article Title</title>}}\\
\hspace{5mm}\small{\texttt{<text>Article Text</text>}}\\
\texttt{\small{<category>Article Categories</category>}}
\end{center}
\vspace{5mm}
The data files are also available in \texttt{plain text, .html, .csv} and \texttt{.json} file formats.

\subsection{Reference Summaries Creation}\label{ref_summary_creation}
Reference summaries are the gold-standard summaries - often created or validated by humans - that serve as benchmarks for evaluating system summaries. In this work, two categories were used: a) the Wikipedia summaries extracted using the Wikipedia API during the text collection stage and b) the summaries created by the human participants. A total of 19 undergraduate and postgraduate students from Cardiff University were recruited to create, summarise and evaluate the articles, 13 of them were undertaking an undergraduate or postgraduate degree in Welsh, which involved previous training on creating summaries from complex texts. The remaining six students were undergraduate students on other degree programmes in Humanities and Social Sciences at Cardiff University and had completed their compulsory education at Welsh-medium or bilingual schools. Students were asked to complete a questionnaire prior to starting work, which elicited biographical information. A total of 17 students had acquired Welsh in the home. One student acquired the language via Welsh-medium immersion education and one student had learned the language as an adult. The majority of students came from south-west Wales (\texttt{n}=11). This region included the counties of Carmarthenshire, Ceredigion, Neath Port Talbot, and Swansea. A further five students came from north-west Wales, which comprised the counties of Anglesey and Gwynedd. One student came from south-east Wales (Cardiff), one from mid Wales (Powys), and one from north-east Wales (Conwy). A broad distinction can be made between northern and southern Welsh. The two varieties (within which further dialectal differences exist) exhibit some differences at all levels of language structure although all varieties are mutually intelligible. Students were asked four questions which elicited information on the lexical, grammatical, and phonological variants they would ordinarily use. The results largely corresponded to geographical area: 11 students used southern forms and seven students used northern forms (including the student from mid Wales). One student, from Cardiff, used a mixture of both northern and southern forms. Students were given oral and written instructions on how to complete the task. Specifically, they were told that the aim of the task was to produce a simple summary for each of the Wikipedia articles (allocated to them) which contained the most important information. They were also asked to conform to the following principles:

\begin{itemize}
\item The length of each summary should be 230 - 250 words.
\item The summary should be written in the author’s own words and not be extracted (copy-pasted) from the Wikipedia article.
\item The summary should not include any information that is not contained in the article
\item Any reference to a living person in the article should be anonymised in the summary (to conform to the ethical requirements of each partner institution).
\item All summaries should be proofread and checked using spell checker software (Cysill) prior to submission\footnote{Cysill: \url{www.cysgliad.com/cy/cysill}}.
\end{itemize}

\begin{figure}[h]
\centering
\includegraphics[scale=0.55]{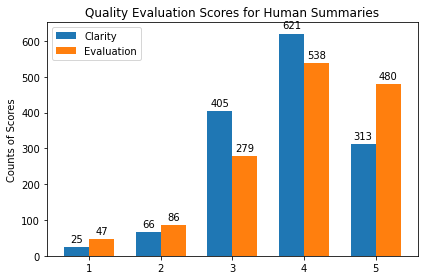}
\caption{Distribution of the readability (clarity) and overall quality evaluation scores for all the 1430 currently available in the Welsh Summarisation Dataset}
\label{fig:qual_eval_human_summ}
\end{figure}

Further instruction was given on the register to be used in the creation of summaries. Students were asked to broadly conform to the principles of Cymraeg Clir \cite{Williams1999} and, in particular, avoid less common short forms of verbs and the passive mode, and use simple vocabulary where possible instead of specialised terms.
Each student completed between 60 - 100 summaries between July and October 2021. The median amount of time spent on each summary was 30 minutes. The complete dataset comprises 1,461 summaries with the remaining 39 summaries not being completed due to one student prematurely dropping out of the project and some instances of unsuitable articles (e.g. lists of bullet points).
Three of the postgraduate students recruited were also asked to evaluate the summaries by giving a score between one and five. Table \ref{tab:markingCriteria} shows the marking criteria.

\begin{table}[ht]
\centering
\begin{tabular}{|c|p{0.4\textwidth}|}
    \hline
        \textbf{Score} &  \textbf{Criteria} \\
         \hline
  \vtop{\hbox{\hbox{\strut}\strut}\hbox{\strut}\hbox{\strut}\hbox{\strut 5}} & \begin{itemize}[leftmargin=*]
  \itemsep-0.5em 
             \item Very clear expression and very readable style. 
             \item Very few language errors.
             \item Relevant knowledge and a good understanding of the article; without significant gaps.
         \end{itemize}\\
                  \hline
  \vtop{\hbox{\hbox{\strut}\strut}\hbox{\strut}\hbox{\strut}\hbox{\strut 4}} &   \begin{itemize}[leftmargin=*]
\itemsep-0.5em
             \item Clear expression and legible style. 
             \item Small number of language errors. 
             \item Relevant knowledge and a good understanding of the article, with some gaps.
         \end{itemize}\\  
        \hline
   \vtop{\hbox{\hbox{\strut}\strut}\hbox{\strut}\hbox{\strut}\hbox{\strut 3}} &   \begin{itemize}[leftmargin=*]
  \itemsep-0.5em
     \item Generally clear expression, and legible style.
     \item Number of language errors. 
     \item The knowledge and understanding of the article is sufficient, although there are several omissions and several errors.
        \end{itemize}\\  
             \hline
   \vtop{\hbox{\hbox{\strut}\strut}\hbox{\strut}\hbox{\strut}\hbox{\strut 2}} &   \begin{itemize}[leftmargin=*]
  \itemsep-0.5em
     \item Expression is generally clear but sometimes unclear. 
     \item Significant number of language errors.
     \item The knowledge and understanding of the article is sufficient for an elementary summary, but there are a number of omissions and errors.
        \end{itemize}\\     
             \hline
    \vtop{\hbox{\hbox{\strut}\strut}\hbox{\strut}\hbox{\strut}\hbox{\strut 1}} &   \begin{itemize}[leftmargin=*]
  \itemsep-0.5em
     \item Expression is often difficult to understand. Defective style.
     \item Persistently serious language errors. 
     \item The information is inadequate for summary purposes. Obvious deficiencies in understanding the article. 
        \end{itemize}\\   
        \hline
\end{tabular}
\caption{Criteria for the marking of summaries}
\label{tab:markingCriteria}
\end{table}

Both the mean and median scores for the summaries were 4. Evaluators were instructed to fix common language errors (such as mutation errors and spelling mistakes) but not to correct syntax. All the participants were duly paid an approved legal wage for their work.

\subsection{Building Summariser Systems}\label{summSys}
The second phase of this summarisation project is to use the corpus dataset to inform the iterative development and evaluation of digital summarisation tools. The main approaches to text summarisation include extraction-based summarisation and abstraction-based summarisation. The former extracts specific words/phrases from the text in the creation of the summary, while the latter works to provide paraphrased summaries (i.e. not directly extracted) from the source text. The successful extraction/abstraction of content, when using summarisation tools/approaches, depends on the accuracy of automatic algorithms (which require training using hand-coded gold-standard datasets).
As an under-resourced language with limited literature on Welsh summarisation, applying summarisation techniques from the literature helps in having initial results that can be used to benchmark the performance of other summarisers on the Welsh language. In this project, we implemented and evaluated basic baseline single-document extractive summarisation systems.

\subsection{Baselines}\label{baselines}
The sections below provide an overview of the summarisation systems that this project will be focusing on currently as well as throughout the life of the project.

\subsubsection{First Sentence Summariser}\label{firstSent}
Rather than using a document's title or keywords \cite{mbonu2021igbosum}, some summarisers tend to use the first sentence of an article to identify the topic to be summarised. The justification behind selecting the first sentence as being representative of the relevant topic is based on the belief that in many cases, especially in news articles or articles found on Wikipedia, the first sentence tends to contain key information about the content of the entire article \cite{radev2004centroid,fattah2008automatic,yeh2008ispreadrank}.

\subsubsection{TextRank}\label{textrank}
This summarisation technique was introduced by \newcite{radev2004centroid}. This was the first graph-based automated text summarisation algorithm that is based on the simple application of the PageRank algorithm. PageRank is used by Google Search to rank web pages in their search engine results \cite{brin1998anatomy}. TextRank utilises this feature to identify the most important sentences in an article.

\subsubsection{LexRank}\label{lexrank}
Similar to TextRank , LexRank uses a graph-based algorithm for automated text summarisation \cite{erkan2004lexrank}. The technique is based on the fact that a cluster of documents can be viewed as a network of sentences that are related to each other. Some sentences are more similar to each other while some others may share only a little information with the rest of the sentences. Like TextRank, LexRank also uses the PageRank algorithm for extracting top keywords. The key difference between the two baselines is the weighting function used for assigning weights to the edges of the graph. While TextRank simply assumes all weights to be unit weights and computes ranks like a typical PageRank execution, LexRank uses degrees of similarity between words and phrases and computes the centrality of the sentences to assign weights \cite{erkan2004lexrank}.

\subsection{Toplines}\label{toplines}
As the project progresses, we will develop more complex summarisers and evaluate their performance by comparing the summarisation results of the three baselines mentioned above. The purpose of the topline summarisers is to prove that using language related technology to summarise Welsh documents will improve the results of those produced by the baseline summarisers.

\subsubsection{TFIDF Summariser}\label{welshtfidf}
Term Frequency Inverse Document Frequency (TFIDF) summarisers work by finding words that have the highest ratio of those words frequency in the document and comparing this rate to their occurrence in the full set of documents to be summarised \cite{salton1983introduction}. TFIDF is a simple numerical statistic which reflect the importance of a word to a document in a text collection or corpus and is usually used as a weighing factor in information retrieval, thus using it to find important sentences in extractive summarisation \cite{Hajime2000comparison,wolf2004summarizing}.
The summariser focuses on finding key and important words in the documents to be summarised in an attempt to produce relevant summaries. Using TFIDF in the Welsh language is not new. \newcite{arthur2019human}, used a social network that they built using Twitter’s geo-locations to identify contiguous geographical regions and identify patterns of communication within and between them. Similarly, we will use TFIDF to identify important sentences based on patterns detected between the summarised document and the summaries corpus.

\subsubsection{TFIDF + Word Embedding}
Here, we used pre-trained word embeddings of features extracted with TFIDF features. The Welsh pre-trained FastText embedding \cite{joulin2016fasttext} which was earlier leveraged by \newcite{ezeani2019leveraging} to fine-tune models for multi-task classification of Welsh part of speech and semantic tagging. \texttt{FastText} extends the \texttt{word2vec} \cite{mikolov2013efficient} approach by substituting words with character n-grams, thereby capturing meanings for shorter words, understanding suffixes and prefixes as well as unknown words.
\\
The experiment was repeated using the \texttt{WNLT} Welsh embeddings by \newcite{corcoran2021creating} who used word2vec and FastText, to automatically learn Welsh word embeddings taking into account syntactic and morphological idiosyncrasies of this language. We will attempt to build upon those two previous efforts enhance the performance of the TFIDF summariser in Section \ref{welshtfidf}.

\subsection{Evaluation} \label{evaluation}
The performance evaluation of the system summarisers was carried out using variants of the ROUGE\footnote{Recall-Oriented Understudy for Gisting Evaluation \cite{lin2004rouge}} metrics. ROUGE measures the quality of the system generated summaries as compared with the reference summaries created or validated by humans (see Section \ref{ref_summary_creation}). The current work uses the ROUGE variants that are commonly applied in literature: \textit{ROUGE-N} (where N= 1 or 2) which considers N-gram text units i.e. unigrams and bigrams; \textit{ROUGE-L} which measures the longest common subsequence in both system and reference summaries while maintaining the order of words; and \textit{ROUGE-SU} is an extended version of \textit{ROUGE-S}\footnote{Default \textit{ROUGE-S} uses skip-gram co-occurrence which considers any pair of words in a sentence allowing for arbitrary gaps while maintaining the order 
} that includes unigrams.\\

Common implementations of ROUGE \cite{ganesan2018rouge} typically produce three key metric scores precision, recall and F1-score as described below.

\[precision = \frac{count(overlapping\ units)}{count(system\ summary\ units)}\]
\[recall = \frac{count(overlapping\ units)}{count(reference\ summary\ units)}\]
\[f1 = (1+\beta^2) * \frac{recall * precision}{recall + \beta^2precision}\]
where the value of \(\beta\) is used to control the relative importance of \textit{precision} and \textit{recall}. Larger \(\beta\) values give more weight to \textit{recall} while \(\beta\) values less than 1 give preference to \textit{precision}. In the current work, \(\beta\) is set to 1 making it equivalent to the harmonic mean between \textit{precision} and \textit{recall}. The term `\textit{units}' as used in the equation refers to either words or n-grams.\\

It is possible to achieve very high recall or precision scores if the system generates a lot more or fewer words than in the reference summary respectively. While we can mitigate that with F1 score to achieve a more reliable measure, we designed our evaluation scheme to investigate the effect of the summary sizes on the performance of the systems. We achieved this by varying the lengths of the system-reference summary pairs during evaluation with \texttt{tokens} = [\texttt{50}, \texttt{100}, \texttt{150}, \texttt{200}, \texttt{250} and \texttt{None}] where \texttt{tokens} indicates the maximum tokens included in the summary and {None} signifies using all the summary at it is. All reported scores are averages of the individual document scores over all the 513 Wikipedia documents used in the experiment.

\begin{figure}[h]
\centering
\includegraphics[scale=0.5]{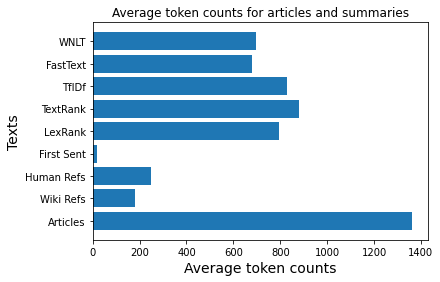}
\caption{Average token counts of the outputs of the systems implemented. This figure shows that given our initial summary size of 50\% of the original article, the outputs of the summariser systems were considerably larger than the reference summaries which explains why we have high recall scores overall.}
\label{fig:average-token-counts}
\end{figure}

\section{Results and Discussion}

\begin{figure*}[htp]
\centering
\includegraphics[scale=0.40]{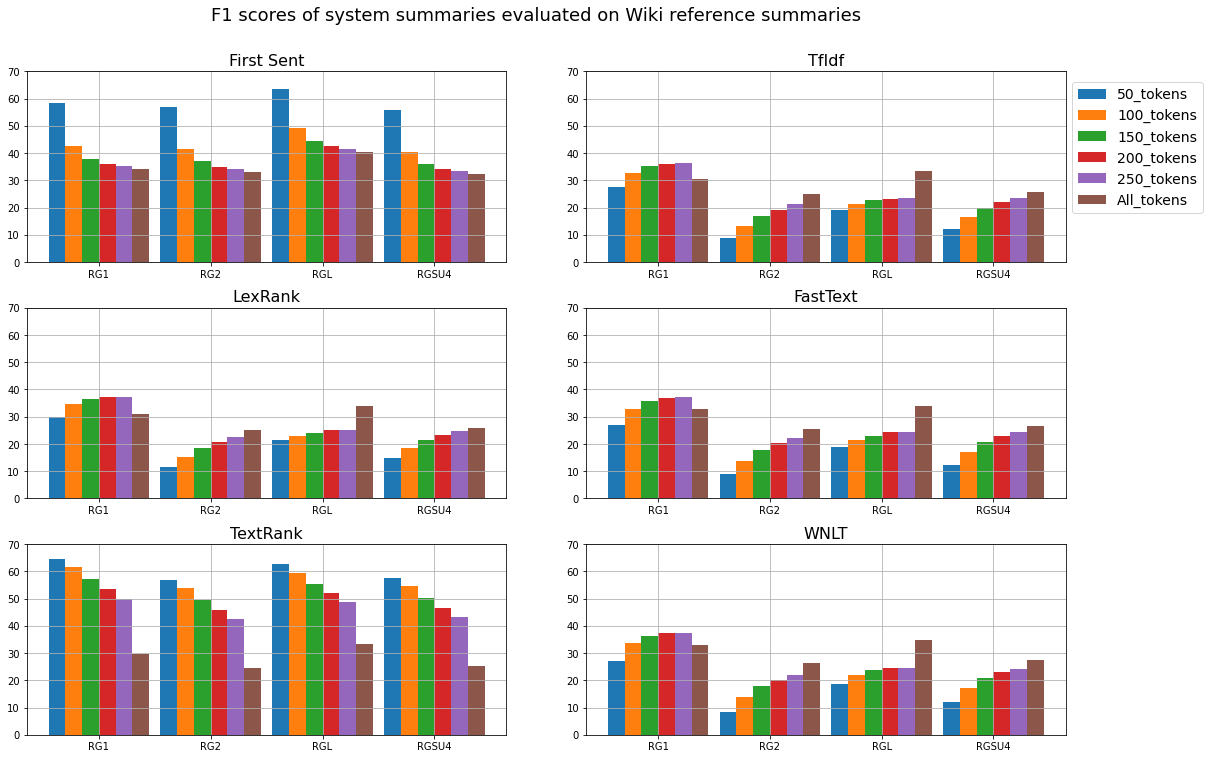}
\includegraphics[scale=0.40]{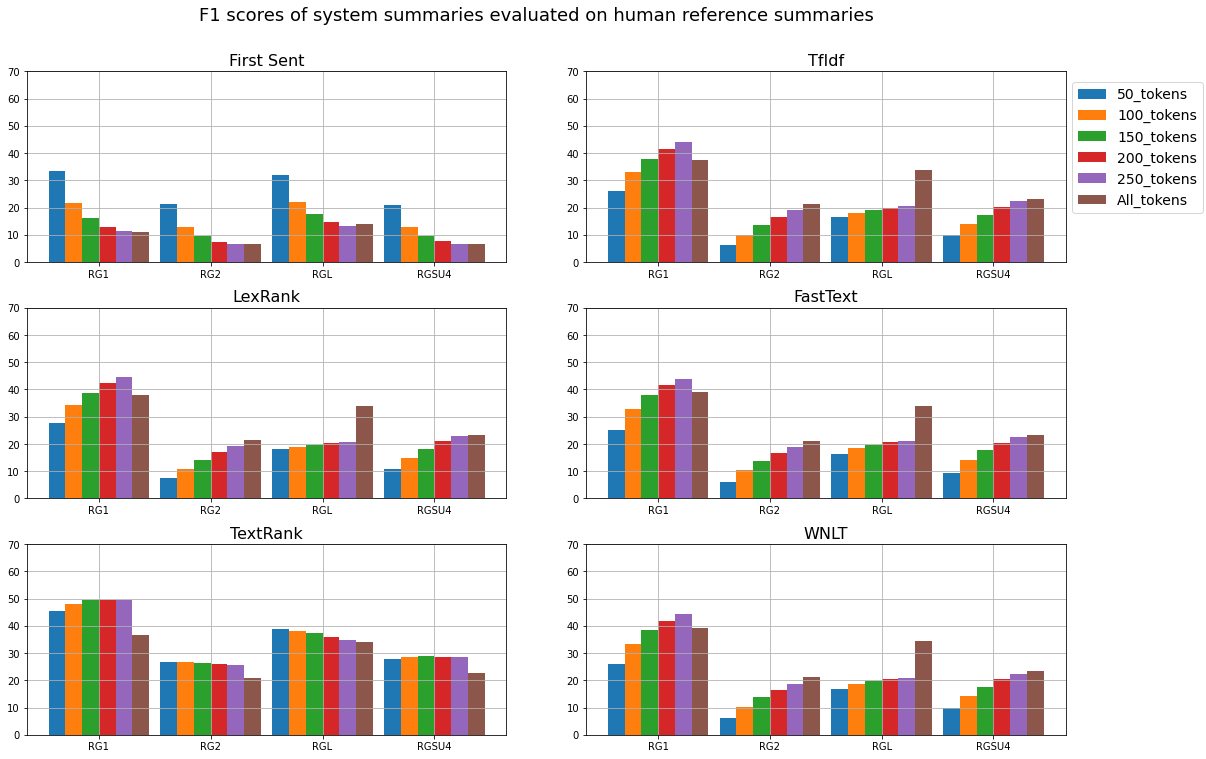}
\caption{F1 scores of system summaries evaluated on Wiki and human reference summaries}
\label{fig:f1_scores}
\end{figure*}

\begin{table*}[h]
    \centering
    \begin{tabular}{|c|c|c|c|c|c||c|c|c|c|}
    \cline{3-10}
    \multicolumn{2}{c|}{}&\multicolumn{4}{|c||}{\textbf{Wiki Refs}} & \multicolumn{4}{c|}{\textbf{Human Refs}} \\
    \cline{3-10}
    \multicolumn{2}{c|}{}&\textbf{RGE-1} & \textbf{RGE-2} & \textbf{RGE-L} & \textbf{RGE-SU4} & \textbf{RGE-1} & \textbf{RGE-2} & \textbf{RGE-L} & \textbf{RGE-SU4}\\
    \hline
\hline
\multirow{3}{*}{1stSent250}
& pre & 99.51 & 99.50 & 99.53 & 99.48 & 70.44 & 42.15 & 58.10 & 44.69\\
& rec & 25.08 & 24.40 & 29.86 & 23.70 & 06.49 & 03.73 & 07.87 & 03.77\\
& f1 & 35.15 & 34.22 & 41.57 & 33.24 & 11.38 & 06.56 & 13.28 & 06.64\\
\hline
\multirow{3}{*}{TextRank250}
& pre & 42.10 & 36.17 & 42.16 & 36.75 & 48.55 & 25.02 & 34.64 & 27.81\\
& rec & 76.23 & 63.10 & 67.90 & 64.89 & 53.45 & 27.48 & 36.68 & 30.56\\
& f1 & \textbf{49.83} & \textbf{42.45} & \textbf{48.88} & \textbf{43.21} & \textbf{49.69} & \textbf{25.57} & \textbf{34.70} & \textbf{28.43}\\
\hline
\multirow{3}{*}{LexRank250}
& pre & 31.50 & 19.07 & 21.40 & 20.94 & 44.14 & 19.25 & 20.91 & 22.80\\
& rec & 58.69 & 34.83 & 37.46 & 38.55 & 47.72 & 20.64 & 21.64 & 24.53\\
& f1 & 37.42 & 22.50 & 25.27 & 24.74 & 44.68 & 19.37 & 20.67 & 23.00\\
\hline
\multirow{3}{*}{TfIDf250}
& pre & 30.57 & 18.00 & 19.77 & 19.94 & 43.34 & 18.75 & 20.46 & 22.32\\
& rec & 56.99 & 32.42 & 35.73 & 36.39 & 47.04 & 20.18 & 21.57 & 24.11\\
& f1 & 36.31 & 21.20 & 23.62 & 23.53 & 43.97 & 18.92 & 20.39 & 22.58\\
\hline
\multirow{3}{*}{FastText250}
& pre & 31.57 & 18.97 & 20.56 & 20.88 & 44.16 & 19.29 & 21.14 & 22.80\\
& rec & 57.66 & 33.00 & 36.58 & 37.00 & 46.65 & 19.98 & 22.15 & 23.81\\
& f1 & 37.18 & 22.01 & 24.42 & 24.31 & 44.00 & 18.95 & 20.91 & 22.52\\
\hline
\multirow{3}{*}{WNLT250}
& pre & 32.03 & 19.01 & 20.87 & 20.97 & 44.69 & 19.29 & 21.30 & 22.93\\
& rec & 57.65 & 32.15 & 36.57 & 36.22 & 46.86 & 19.73 & 22.13 & 23.71\\
& f1 & 37.50 & 21.82 & 24.65 & 24.19 & 44.28 & 18.76 & 20.89 & 22.48\\
\hline\hline
\multirow{3}{*}{1stSent}
& pre & 99.51 & 99.50 & 99.53 & 99.48 & 70.52 & 42.20 & 61.69 & 44.62\\
& rec & 24.45 & 23.79 & 29.03 & 23.11 & 06.34 & 03.71 & 08.25 & 03.77\\
& f1 & 34.07 & 33.17 & 40.26 & 32.23 & 11.15 & 06.53 & 13.94 & 06.65\\
\hline
\multirow{3}{*}{TextRank}
& pre & 21.12 & 17.98 & 24.47 & 18.67 & 27.60 & 15.90 & 26.85 & 17.36\\
& rec & 81.91 & 64.62 & 73.78 & 66.19 & 70.17 & 39.89 & 56.05 & 42.82\\
& f1 & 29.56 & 24.61 & 33.28 & 25.40 & 36.73 & 21.04 & 33.97 & 22.83\\
\hline
\multirow{3}{*}{LexRank}
& pre & 22.90 & 19.04 & 25.57 & 19.88 & 30.11 & 17.09 & 27.86 & 18.79\\
& rec & 79.32 & 60.95 & 70.82 & 62.54 & 67.46 & 37.54 & 53.10 & 40.43\\
& f1 & 30.98 & 25.07 & 33.81 & 25.98 & 38.12 & \textbf{21.43} & 33.97 & 23.39\\
\hline
\multirow{3}{*}{TfIdf}
& pre & 22.25 & 18.66 & 24.92 & 19.42 & 29.01 & 16.52 & 27.16 & 18.13\\
& rec & 81.06 & 62.80 & 72.75 & 64.38 & 68.81 & 38.39 & 54.68 & 41.37\\
& f1 & 30.52 & 24.95 & 33.49 & 25.80 & 37.56 & 21.20 & 33.83 & 23.12\\
\hline
\multirow{3}{*}{FastText}
& pre & 25.03 & 20.27 & 26.31 & 21.30 & 32.71 & 17.88 & 28.90 & 19.97\\
& rec & 76.14 & 55.87 & 67.26 & 57.90 & 64.24 & 33.95 & 50.57 & 37.23\\
& f1 & 32.65 & 25.46 & 33.95 & 26.57 & 39.26 & 21.06 & 33.92 & 23.38\\
\hline
\multirow{3}{*}{WNLT}
& pre & 25.25 & 20.88 & 26.94 & 21.83 & 32.60 & 18.06 & 29.14 & 20.07\\
& rec & 78.01 & 58.04 & 69.56 & 60.01 & 65.61 & 35.18 & 51.93 & 38.44\\
& f1 & \textbf{33.13} & \textbf{26.43} & \textbf{34.95} & \textbf{27.46} & \textbf{39.28} & 21.33 & \textbf{34.29} & \textbf{23.58}\\
\hline
    \end{tabular}
    \caption{Results of evaluating \textbf{Baseline} (\textit{First sentence} (Bottomline), \textit{LexRank}, \textit{TextRank}) and \textbf{Topline} (\textit{TfIdf}, \textit{Fasttext} and \textit{WNLT} word embedding) model summaries on the combined Wiki and Human reference summaries}
    \label{tab:eval_results}
\end{table*}

Figure \ref{fig:f1_scores} shows the plots of the ROUGE metric f1 scores for all the system summaries evaluated on the reference summaries. Each bar represents the score for a different maximum length setting - 50, 100, 150, 200, 250 and None - as described in Section \ref{evaluation}. Table \ref{tab:eval_results} shows the full metric scores for only the last set of scores (i.e. 250 and None) due to space constraints.
\\ \\
Table \ref{tab:eval_results} and Figure \ref{fig:f1_scores} show the summary of our initial experiments and evaluations of the system summaries on both the Wikipedia and human summaries. Decent results were achieved across the systems even with short summaries. In particular, Figure \ref{fig:f1_scores} shows that \texttt{TextRank}'s scores improves with fewer tokens achieving the best overall score on the controlled token length evaluations. However, its overall scores drop as the length of the summaries increase.
\\ \\
The plots clearly show that there is a performance improvement between from the bottom line model, \texttt{First Sent}, to the topline models. The high precision score from \texttt{First Sent} could be explained by the fact that some of Wikipedia summaries are often generated using similar automatic techniques. But its comparatively low recall scores would be because as shown in Figure \ref{fig:average-token-counts} the reference summaries it is evaluated are significantly larger than its summaries which are made up of only one sentence - the first sentence of the article. The other systems however returned higher recall scores because, compared to the system summaries, the reference summaries were significantly smaller.

Another key point on from Figure \ref{fig:f1_scores} is the similarity in the plots of the TFIDF based systems as well as \texttt{LexRank}. It appears that the cosine-similarity score, which is the underlying measure for the ranking algorithm shared among, has a major impact in how they work. It is also interesting that while \texttt{TextRank}'s scores dropped as the size of the summary increases, the reverse is the case for the others. There is a general drop in performance on the human summaries when compared with the Wiki summaries. This is a confirmation that despite the good results generated by the system, they still could not match the inherent qualities - coherence, consistency, fluency and relevance - embedded in human created summaries. As mentioned in Section \ref{conclusion}, building and deploying Welsh summarisers - extractive and abstractive - based on the state-of-the-art transformer models is the current of focus of this work.
\\ \\
Overall, discounting the \texttt{First Sent} scores, the TFIDF+embedding based models gave the best f1 scores on summaries on longer summaries while \texttt{TextRank} consistently outperformed the others systems on shorter summaries.

\section{Conclusion and future work} \label{conclusion}
This work presents a new publicly available and freely accessible high-quality Welsh text summarisation dataset as well as the implementation of basic extractive summarisation systems. Given that Welsh is considered low-resourced with regards to NLP, this dataset will enable further research works in Welsh automatic text summarisation systems as well as Welsh language technology in general. Overall, the development of the automated tools for Welsh language and facilitate the work of those involved in document preparation, proof-reading, and (in certain circumstances) translation. 
\\\\
We are currently focusing on leveraging the existing state-of-the-art transformer based models for building and deploying Welsh text summariser model. The summarisation state of the art literature shows a great shift towards using deep learning to create extractive and abstractive supervised and unsupervised summarisers using deep learning models such as Convolutional Neural Network (CNN), Recurrent Neural Network (RNN), Long Short Term Memory (LSTM) and many others \cite{song2019abstractive,zmandar2021financial,zmandar2021joint,magdum2021survey}. In this project we will combine the use of the aforementioned Welsh word embeddings to try and improve the results and create Welsh summarisation systems that are on par with other English and European state of the art summarisers.
\\\\
The Welsh summariser tool will allow professionals to quickly summarise long documents for efficient presentation. For instance, the tool will allow educators to adapt long documents for use in the classroom. It is also envisaged that the tool will benefit the wider public, who may prefer to read a summary of complex information presented on the internet or who may have difficulties reading translated versions of information on websites. To keep up to date with developments on this tool, please visit the main project website\footnote{\url{https://corcencc.org/resources/\#ACC}}. 

\section{Acknowledgements}
This research was funded by the Welsh Government, under the Grant `Welsh Automatic Text Summarisation'. We are grateful to Jason Evans, National Wikimedian at the National Library of Wales, for this initial advice.

\section{Bibliographical References}\label{reference}

\bibliographystyle{lrec2022-bib}
\bibliography{lrec2022}

\begin{thebibliography}{}

\bibitem[\protect\citename{Arthur and Williams}2019]{arthur2019human}
Arthur, R. and Williams, H.~T.
\newblock (2019).
\newblock The human geography of twitter: Quantifying regional identity and
  inter-region communication in england and wales.
\newblock {\em PloS one}, 14(4):e0214466.

\bibitem[\protect\citename{Brin and Page}1998]{brin1998anatomy}
Brin, S. and Page, L.
\newblock (1998).
\newblock The anatomy of a large-scale hypertextual web search engine.
\newblock {\em Computer networks and ISDN systems}, 30(1-7):107--117.

\bibitem[\protect\citename{Carlin and Chr{\'\i}ost}2016]{carlin2016standard}
Carlin, P. and Chr{\'\i}ost, D. M.~G.
\newblock (2016).
\newblock A standard for language? policy, territory, and constitutionality in
  a devolving wales.
\newblock In {\em Sociolinguistics in Wales}, pages 93--119. Springer.

\bibitem[\protect\citename{Corcoran \bgroup et al.\egroup
  }2021]{corcoran2021creating}
Corcoran, P., Palmer, G., Arman, L., Knight, D., and Spasi{\'c}, I.
\newblock (2021).
\newblock Creating welsh language word embeddings.
\newblock {\em Applied Sciences}, 11(15):6896.

\bibitem[\protect\citename{Cunliffe \bgroup et al.\egroup
  }2013]{cunliffe2013young}
Cunliffe, D., Morris, D., and Prys, C.
\newblock (2013).
\newblock Young bilinguals' language behaviour in social networking sites: The
  use of welsh on facebook.
\newblock {\em Journal of Computer-Mediated Communication}, 18(3):339--361.

\bibitem[\protect\citename{El-Haj and Rayson}2013]{el2013using}
El-Haj, M. and Rayson, P.
\newblock (2013).
\newblock Using a keyness metric for single and multi document summarisation.
\newblock In {\em Proceedings of the MultiLing 2013 Workshop on Multilingual
  Multi-document Summarization}, pages 64--71.

\bibitem[\protect\citename{El-Haj \bgroup et al.\egroup }2011]{el2011multi}
El-Haj, M., Kruschwitz, U., and Fox, C.
\newblock (2011).
\newblock Multi-document arabic text summarisation.
\newblock In {\em 2011 3rd Computer Science and Electronic Engineering
  Conference (CEEC)}, pages 40--44. IEEE.

\bibitem[\protect\citename{Erkan and Radev}2004]{erkan2004lexrank}
Erkan, G. and Radev, D.~R.
\newblock (2004).
\newblock Lexrank: Graph-based lexical centrality as salience in text
  summarization.
\newblock {\em Journal of artificial intelligence research}, 22:457--479.

\bibitem[\protect\citename{Ezeani \bgroup et al.\egroup
  }2019]{ezeani2019leveraging}
Ezeani, I., Piao, S.~S., Neale, S., Rayson, P., and Knight, D.
\newblock (2019).
\newblock Leveraging pre-trained embeddings for welsh taggers.
\newblock In {\em Proceedings of the 4th Workshop on Representation Learning
  for NLP (RepL4NLP-2019)}, pages 270--280.

\bibitem[\protect\citename{Fattah and Ren}2008]{fattah2008automatic}
Fattah, M.~A. and Ren, F.
\newblock (2008).
\newblock Automatic text summarization.
\newblock {\em World Academy of Science, Engineering and Technology},
  37(2):192.

\bibitem[\protect\citename{Ganesan}2018]{ganesan2018rouge}
Ganesan, K.
\newblock (2018).
\newblock Rouge 2.0: Updated and improved measures for evaluation of
  summarization tasks.

\bibitem[\protect\citename{Goldstein \bgroup et al.\egroup
  }2000]{goldstein2000multi}
Goldstein, J., Mittal, V.~O., Carbonell, J.~G., and Kantrowitz, M.
\newblock (2000).
\newblock Multi-document summarization by sentence extraction.
\newblock In {\em NAACL-ANLP 2000 Workshop: Automatic Summarization}.

\bibitem[\protect\citename{Joulin \bgroup et al.\egroup
  }2016]{joulin2016fasttext}
Joulin, A., Grave, E., Bojanowski, P., Douze, M., J{\'e}gou, H., and Mikolov,
  T.
\newblock (2016).
\newblock Fasttext. zip: Compressing text classification models.
\newblock {\em arXiv preprint arXiv:1612.03651}.

\bibitem[\protect\citename{Lin}2004]{lin2004rouge}
Lin, C.-Y.
\newblock (2004).
\newblock Rouge: A package for automatic evaluation of summaries.
\newblock In {\em Text summarization branches out}, pages 74--81.

\bibitem[\protect\citename{Litvak and Last}2008]{litvak2008graph}
Litvak, M. and Last, M.
\newblock (2008).
\newblock Graph-based keyword extraction for single-document summarization.
\newblock In {\em Coling 2008: Proceedings of the workshop Multi-source
  Multilingual Information Extraction and Summarization}, pages 17--24.

\bibitem[\protect\citename{Magdum and Rathi}2021]{magdum2021survey}
Magdum, P. and Rathi, S.
\newblock (2021).
\newblock A survey on deep learning-based automatic text summarization models.
\newblock In {\em Advances in Artificial Intelligence and Data Engineering},
  pages 377--392. Springer.

\bibitem[\protect\citename{Mbonu \bgroup et al.\egroup }2021]{mbonu2021igbosum}
Mbonu, C., Chukwuneke, C.~I., Paul, R., Ezeani, I., and Onyenwe, I.
\newblock (2021).
\newblock {IgboSum}1500 - {Introducing} the {Igbo Text Summarization Dataset}.
\newblock In {\em 3rd Workshop on African Natural Language Processing}.

\bibitem[\protect\citename{Mikolov \bgroup et al.\egroup
  }2013]{mikolov2013efficient}
Mikolov, T., Chen, K., Corrado, G., and Dean, J.
\newblock (2013).
\newblock Efficient estimation of word representations in vector space.
\newblock {\em arXiv preprint arXiv:1301.3781}.

\bibitem[\protect\citename{Mochizuki and Okumura}2000]{Hajime2000comparison}
Mochizuki, H. and Okumura, M.
\newblock (2000).
\newblock A comparison of summarization methods based on task-based evaluation.
\newblock In {\em LREC}.

\bibitem[\protect\citename{Radev \bgroup et al.\egroup
  }2004]{radev2004centroid}
Radev, D.~R., Jing, H., Sty{\'s}, M., and Tam, D.
\newblock (2004).
\newblock Centroid-based summarization of multiple documents.
\newblock {\em Information Processing \& Management}, 40(6):919--938.

\bibitem[\protect\citename{Salton and McGill}1983]{salton1983introduction}
Salton, G. and McGill, M.~J.
\newblock (1983).
\newblock {\em Introduction to modern information retrieval}.
\newblock mcgraw-hill.

\bibitem[\protect\citename{Song \bgroup et al.\egroup
  }2019]{song2019abstractive}
Song, S., Huang, H., and Ruan, T.
\newblock (2019).
\newblock Abstractive text summarization using lstm-cnn based deep learning.
\newblock {\em Multimedia Tools and Applications}, 78(1):857--875.

\bibitem[\protect\citename{Svore \bgroup et al.\egroup
  }2007]{svore2007enhancing}
Svore, K., Vanderwende, L., and Burges, C.
\newblock (2007).
\newblock Enhancing single-document summarization by combining ranknet and
  third-party sources.
\newblock In {\em Proceedings of the 2007 joint conference on empirical methods
  in natural language processing and computational natural language learning
  (EMNLP-CoNLL)}, pages 448--457.

\bibitem[\protect\citename{Williams}1999]{Williams1999}
Williams, C.
\newblock (1999).
\newblock {\em Cymraeg Clir: Canllawiau Iaith}.
\newblock Bangor: Gwynedd Council, Welsh Language Board and Canolfan Bedwyr.

\bibitem[\protect\citename{Wolf \bgroup et al.\egroup
  }2004]{wolf2004summarizing}
Wolf, C.~G., Alpert, S.~R., Vergo, J., Kozakov, L., and Doganata, Y.
\newblock (2004).
\newblock Summarizing technical support documents for search: expert and user
  studies.
\newblock {\em IBM systems journal}, 43(3):564--586.

\bibitem[\protect\citename{Yeh \bgroup et al.\egroup }2008]{yeh2008ispreadrank}
Yeh, J.-Y., Ke, H.-R., and Yang, W.-P.
\newblock (2008).
\newblock ispreadrank: Ranking sentences for extraction-based summarization
  using feature weight propagation in the sentence similarity network.
\newblock {\em Expert Systems with Applications}, 35(3):1451--1462.

\bibitem[\protect\citename{Zmandar \bgroup et al.\egroup
  }2021a]{zmandar2021financial}
Zmandar, N., El-Haj, M., Rayson, P., Litvak, M., Giannakopoulos, G., Pittaras,
  N., et~al.
\newblock (2021a).
\newblock The financial narrative summarisation shared task fns 2021.
\newblock In {\em Proceedings of the 3rd Financial Narrative Processing
  Workshop}, pages 120--125.

\bibitem[\protect\citename{Zmandar \bgroup et al.\egroup
  }2021b]{zmandar2021joint}
Zmandar, N., Singh, A., El-Haj, M., and Rayson, P.
\newblock (2021b).
\newblock Joint abstractive and extractive method for long financial document
  summarization.
\newblock In {\em Proceedings of the 3rd Financial Narrative Processing
  Workshop}, pages 99--105.

\end{thebibliography}

\end{document}